\documentclass{article}
\usepackage{ijcai18}

\usepackage{microtype}
\usepackage{graphicx}
\usepackage{subfigure}
\usepackage{booktabs} 

\usepackage{amsmath}
\usepackage{tabularx}
\usepackage{tabu}
\usepackage{mathtools}
\usepackage{nicefrac}
\usepackage{natbib}
\usepackage{algorithm}
\usepackage{algorithmic}
\usepackage{times}

\title{Explicating feature contribution using Random Forest proximity distances}
\author{Leanne S. Whitmore, Anthe George, Corey M. Hudson\\ 
Sandia National Laboratories, Livermore, CA 94551  \\
cmhudso@sandia.gov}

\begin{document}

\maketitle

\begin{abstract}
In Random Forests, proximity distances are a metric representation of data into decision space. By observing how changes in input map to the movement of instances in this space we are able to determine the independent contribution of each feature to the decision-making process. For binary feature vectors, this process is fully specified. As these changes in input move particular instances nearer to the in-group or out-group, the independent contribution of each feature can be uncovered. Using this technique, we are able to calculate the contribution of each feature in determining how black-box decisions were made. This allows explication of the decision-making process, audit of the classifier, and {\it post-hoc} analysis of errors in classification.
\end{abstract}

\section{Introduction}

Limitations in the ability to identify the pathway to decision-making for individual instances are the fundamental justification for a classifier being labeled "black box" [\cite{caruana1999case}]. Visibility of the decision-making for individual cases through a machine learned model, what we here refer to as {\it explication}, is essential for algorithmic transparency [\cite{lipton2016mythos}]. One of the most popular "black box" classifiers is the Random Forest Classifier and its associated variants (including Bagging Classifiers and Gradient Boosted Trees Classifiers). Despite the common use of these techniques for machine learning, in high risk or high consequence settings, lower accuracy methods are often preferred, due to their interpretability [\cite{caruana2015intelligible}]. 

The manner in which Random Forests work has been well described by Breiman (\citeyear{breiman2001random}), including how they limit overfitting (using the Strong Law of Large Numbers) and achieve their level of accuracy (through improvements to bagging and boosting algorithms and decorrelation of features in random samples). In spite of Random Forests implementing a straightforward construction algorithm, the intricacies of a model's decision-making processes is largely hidden from the outcome [\cite{friedman2001elements}]. This separates Random Forests from techniques like Logistic Regression Classification, in which individual coefficients correspond directly to weighting factors [\cite{cooper1997evaluation}]. With these more interpretable, but simpler models, running an individual instance through the model, is useful in explaining the decision-making process for that instance. The coefficients in a logistic regression model correspond directly to the odds outcome weightings for each feature. Decisions, therefore can be made, using these features that will directly affect the outcome (e.g., [\cite{herrmann1992serum}]). However, logistic regression is seldom very accurate and in high risk setting, operators are frequently forced to choose between less accurate methods with more explainable models [\cite{bratko1997machine,caruana2017intelligible}].

\subsection{Model feature importances are limited in terms of explicability}

Most of the interpretability in Random Forests come from a set of outcome variables, referred to as feature or variable importances. These values are calculated a number of ways, but are generally measured as the mean decrease in accuracy across the model after traversing tree nodes and splitting on the feature of interest [\cite{friedman2001elements}]. As Breiman and Cutler (\citeyear{breiman2008random}) describe it, this requires a simple random permutation of a variable in the sampled instances, run through the model. The number of correctly classified counts are then compared between the sampled instances (without permutation) and the sampled instances (with permutation). The average over all trees is the feature importance score for that variable.

One of the limitations of feature importances, in terms of interpretability, is that they measure the importance of the feature in the model, and not to individual cases. The manner in which feature importances relate to individual instances is not clear from the feature importance values. Highly weighted features may have little importance for a particular instance. They may also be positive indicators of class membership for one instance and negative indicators of class membership for another instance, even when the outcome prediction is identical. Feature importances give global importance, but not local importance or the direction in which they drive a particular decision-making process. The global role of features importance measurements cannot provide the level of explanation that is frequently found in more causal measures, like interventions [\cite{garant2016evaluating,galles1995testing}].

\subsection{Related work in feature contribution and explicability}

Recently, a variety of techniques, under the general banner of \"counterfactual probing\" techniques have emerged in machine learning [\cite{voosen2017ai,wachter2017counterfactual}]. These techniques attempt to explain and explicate machine learning decision-making processes. In general, the central idea of these techniques, has been to insert randomness into the feature space, in order to determine the resulting outcome in decision space [\cite{lundberg2017unified,vstrumbelj2014explaining}]. One of the most general purpose of these is LIME (Locally Interpretable Machine-Agnostic Explanations), which maps a complicated and uninterpretable model into linear space, to provide local model fidelity and explainability [\cite{ribeiro2016model}]. Other work in this area has involved determining output or internal neural network gradients through input perturbations [\cite{shrikumar2017learning,lengerichtowards}], additive feature coalition detection [\cite{lundberg2017consistent}], and probability series expansion of feature groupings [\cite{agarwal2017probability}]. 

The primary contribution of this paper is to provide work parallel to the above-listed works on explainability/explicability. Rather than creating a new classifier or new metrics for interpreting classifiers, we seek to provide a topological interpretation of proximity (a well known metric) for Random Forests (a commonly used classifier). Using this topological interpretation, we are able to show how the perceived black-box outcome of individual Random Forests decision-making processes can be made explicable through random permutation of given feature values. 

\section{Representing Random Forests in Decision Space}

By default, many implementations of Random Forest classification use the CART algorithm. This  algorithm represents the decision-making process as a set of discrete splits, ultimately creating binary decision trees [\cite{breiman1984classification}]. Within the mapping from feature vectors to each decision tree, the data is re-represented as a discrete binary vector. Determining the contribution of individual features in the Random Forests decision-making process, requires determining how each value in feature space maps an instance in decision space. For classification operations on binary feature vectors, this process for independently described features is fully specifiable - meaning that the full range of variability in binary features can be captured in the model. In many classification and regression tasks, specification of only binary features is an unsuitable abstraction. However, the underlying data structure of Random Forests is the CART decision tree. To accommodate this data structure, categorical and continuous variable are discretized into binary features. For example the single feature $X_i\in{a,b,c}$ is frequently converted into three binary features $X_{ia}\in \{0,1\}$, $X_{ib}\in \{0,1\}$, $X_{ic}\in \{0,1\}$ and $X_i = \{x \in \mathbf{R} \mid 0 \le x \le 1 \}$ is frequently split into discrete parts, determined by the optimal Gini impurity, e.g., $X_{i1} = \{x \in \{0,1\} \mid x < 0.25 \}$, $X_{i2} = \{x \in \{0,1\} \mid x < 0.71 \}$, $X_{i3} = \{x \in \{0,1\} \mid x > 0.5 \}$, in which case a value x = 0.3 will correspond to a feature vector 010 and a value x = 0.65 will correspond to a feature vector 011. This is a fundamental characteristic of tree-based classifiers, and separate this class of machine learning methods from methods that use a maximally discriminative hyperplane, like Support Vector Machines or Artificial Neural Networks. Random Forests make decisions about how to convert numerical features to binary features. Between user input and the creation of binary feature vectors there is a process of choosing how to split numerical variables. One reason to specifically focus on binary feature vectors is to address feature value permutation, rather than the process of turning numerical variables into binary variables. A numerical feature may have values between 0 and 100, but split at 1.1. Changing feature values across the range of values is meaningless if it undersamples between 0 and 1.1 and oversamples between 1.1 and 100. If only two binary features correspond to the numerical value then there are only two feature values that contribute to the predicted outcome.

\subsection{Defining a proximity distance}
\label{proximity}

The determination of proximity between two instances in a Random Forest data set was defined by Breiman (\citeyear{breiman2001random}) as the number of times two instances share a leaf node in a Decision Tree divided by the number of trees in a Random Forest. Proximity is a similarity metric that accounts for the similarity in the outcome of a hidden decision-making process. Since proximity is a dot-product between 0/1 vectors, per the leaf indicator functions, the Euclidean space that Breiman (\citeyear{breiman2001random}) defines is necessarily similar to the Hamming distance, which has previously been shown to satisfy the properties of a metric.

\subsection{Proximity distance defines the Random Forest decision space}

The totality of all $D(x,.)$ for all $x$ in the training set is the training proximity distance space $P$. All instances sharing the same class as a particular instance $x$ can be labeled as an instance's in-group. $P_x(in)$ defines the covering of in-group proximity space. $P_x(out)$ covers all other distances, and can be labeled as the out-group proximity space for instance $x$. Breiman (\citeyear{breiman2002manual}) defined the average proximity for case $n$ to all other training instances $k$, where $n \ne k$ in class $j$ as:
\begin{equation}
	\label{eq:average_proximity}
	\bar{P} = \sum_{n=1}^{\parallel n \in (class(j)) \parallel} proximity(n,k)^2
\end{equation}

This value can be defined both for the in-groups and out-groups. This can be similarly be defined for proximity distance (1 - proximity) with the opposite interpretation. In this case, small average in-group proximity distances correspond to closeness in decision space. Breiman and Cutler (\citeyear{breiman2001random}) point out that this measure can also be used to threshold outliers, for instances where the average in-group proximity distance is large.

\begin{figure}[t]
\begin{center}
\centerline{\includegraphics[width=\columnwidth]{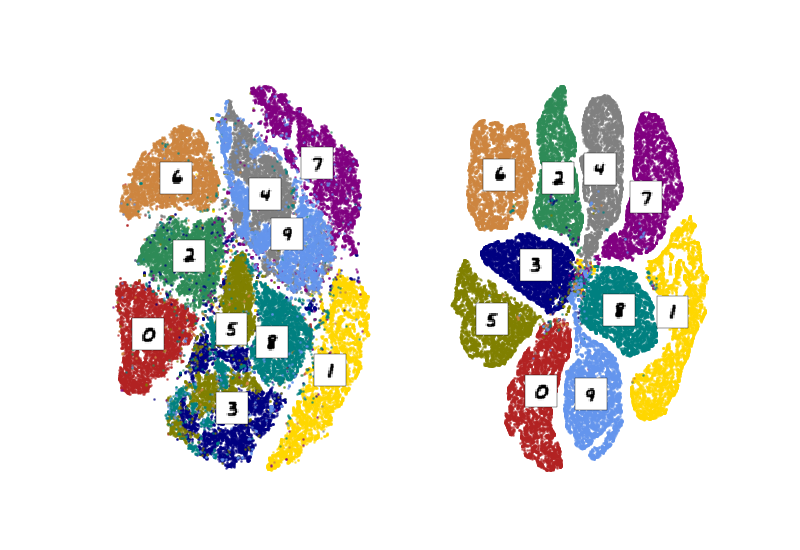}}
\caption{Data are plotted using t-SNE (perplexity = 30, number of iterations = 1000, learning rate = 200). For the distance space figure ({\it left}) the data are Hamming distances of binary feature vectors. For the proximity distance space ({\it right}) the data are transformed using proximity and then mapped into lower dimensional space with t-SNE. Kullback-Leibler divergence for the Hamming distance projection is 3.21 and for the proximity distance projection is 2.96.}
\label{fig:output_coords}
\end{center}
\vskip -0.2in
\end{figure}

The complete matrix of proximity distances defines the proximity space. Distances in decision space, may be quite different than distances in feature space. In feature space, the feature vectors are weighted, either equally, or by {\it a priori} determination. The importance of a value in the input vector, with regard to discriminating the instance into a class is therefore unknown, although, this is frequently a goal of unsupervised learning. Proximity distances transform input features through the decision trees in the Random Forest (see Fig. \ref{fig:output_coords}). As such, sharing few highly discriminating features may be more important for placing two instances in a leaf node than sharing many more features with less discriminating power. In many ways, this is a goal in feature salience studies, which seek to reduce the space of features [\cite{wang2001comparative}] - however, our technique uses all provided features, in order to create a proximity space.

\subsection{Using Proximity Distance to explicate feature contribution}
\label{proximity_defined}

Zhou et al. (\citeyear{zhou2010gene}) recognized that change in proximity matrices could be used to identify model feature importances. However, their study of genomic features compared the result of different treatments in the decision-making process. In this work, we seek to individually explain the manner in which features independently contribute to the specific decision-making processes. This is the fundamental difference between feature importance (a property of the model) and feature contribution (a property of the individual decision-making process). As mentioned above, when the data are presented in binary feature vectors, the process of identifying the contribution of specific features in the decision making for individual instances is fully specifiable. 

Each value in an input feature vector has the potential to contribute to the output decision for a particular instance, by driving the decision-making process toward a different leaf node in the decision trees. Changing a value in a feature vector and running them through the decision trees in a Random Forest can have the effect of moving an instance nearer or farther from the in-group and out-group, orthogonal to both, or resulting in no change in $P$. A given feature, $k$, in a binary feature vector $v$, maps to $\varphi(v)$. Bit-wise permutation of the value $k$, in $v$, defined as $v^{\prime}$, maps to $\varphi(v^{\prime})$. Distance in proximity distance space, can be determined by fixing $P$ for the original dataset and determining a new vector of distances for $D(v^{\prime},.)$ for all $x$ in $P$. The change in position between $D(v^{\prime},.)$ and $D(v,.)$ determines the contribution of the value of feature $k$, given identical values for all other features, between $v^{\prime}$  and $v$. 

\begin{figure}[t]
\begin{center}
\centerline{\includegraphics[width=\columnwidth]{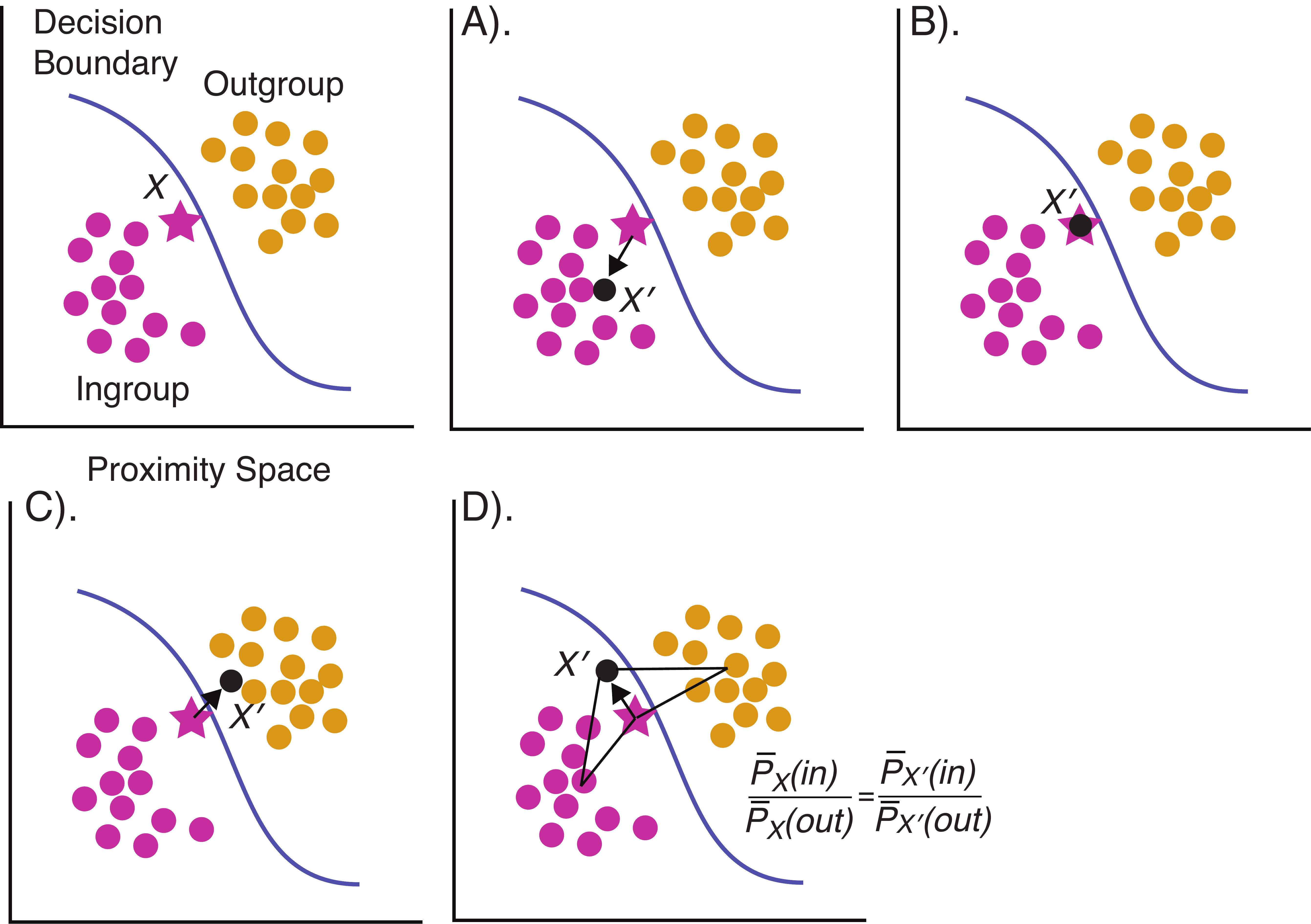}}
\caption{Results of $k$ value permutation. The space ($P$) is defined by all relative distances. The decision boundary here is a simple function. In real decision space, for Random Forests, these are often high-dimensional, discontinuous and non-linear. Relative to $X$ input vector, there is an ingroup and an outgroup. A) A change in $k$, may move $X^{\prime}$ closer to the ingroup. This suggests that $k$ weakened the ultimate decision, proportional to the distance between $X$ and $X^{\prime}$. B) A change in $k$, may have little change in the relative position of $X^{\prime}$ with regard to $X$, suggesting that $k$ had little to no influence on the decision. C) A change in $k$ may result in $X^{\prime}$ being closer to the outgroup. This suggests that $k$ contributed to the ultimate decision, at a strength proportional to the distance between $X$ and $X^{\prime}$. D) Suggests that a change in $k$ resulted in $X^{\prime}$ being proportionally equivalent to both the ingroup and outgroup. If $\bar{P_x}(in) < \bar{P_{x^{\prime}}}(in)$, then $k$ contributed to $X$ being an inlier in decision space. If $\bar{P_x}(in) > \bar{P_{x^{\prime}}}(in)$, then $k$ contributed to $X$ being an outlier in decision space.}
\label{fig:Proximity}
\end{center}
\vskip -0.2in
\end{figure}

The value of $k$ in the original input feature vector may result in one of several potentially overlapping outcomes in decision space. 1) The value may strongly contribute to the decision. 2) The value may weakly impact on the decision. 3) The value may have no impact on the decision. 4) The value may result in the outcome being closer to in-group. 5) The value may result in the outcome being closer to out-groups. The input value may result in a larger average in-group proximity distance, and larger out-group proximity distance (the value is highly discriminating, but moves the instance toward becoming an outlier). The input value may result in a larger average in-group proximity distance and smaller out-group proximity distance (the value moves the instance toward other classes). The input value may also lead to no change in the placement of the instance in the leaf nodes, leading to no change, or strictly orthogonal placement in decision space. A hypothetical set of outcomes in decision space can be visualized in Fig \ref{fig:Proximity}.

\section{Algorithmic Determination of Feature Contribution using k-permutations}

Using the logic of \ref{proximity_defined} as our basis, we here use permuted bit-flipping to determine the contribution of a feature in mapping instance values in decision space. This requires a binary feature vector ($v$), and a Random Forest classifier ($C$) that was trained with $X$. The class of ($v$) is $c$. The $ProximityDistance$ is calculated as the number of features subtracted by the number of shared terminal nodes (between the original and permuted values) divided by the number of features. Each pair of values (original vs. permuted) are then subtracted and squared, giving a distance.

\subsection{Feature contribution and closeness}

There are two main values for each feature in decision space: contribution and closeness. Contribution quantifies the independent influence of each feature on the final position of an instance in proximity distance space. This is determined by modifying each feature value and measuring the distance between the original ($v$) and modified instances ($v^{\prime}$). The measurement of contribution is done by comparing each proximity distance between $v$ and all $v^{\prime}$. Absolute distance is simply calculated as $D(v,v^{\prime})$. Absolute distance, however gives no sense of contribution of the feature to the decision-making process, only the relative number of changes in the word vectors $\varphi(v)$ and $\varphi(v^{\prime})$. 

Feature contribution can be calculated by creating an in-group vector $\mathbf{z}$ relative to the class $c$ of $v$, the size of the number of instances in the training set $X$. In this case 

\begin{equation}
	z_i 
    \begin{cases}
  	-1 & \text{ if } C(v) = C(X_i)\\    
  	 1 & \text{ if } C(v) \neq C(X_i)
	\end{cases}
\end{equation}
Calculating the dot-product of $\mathbf{z}$ by the squared difference between the distance vectors determines the magnitude and direction of the contribution of the feature changed in $v^{\prime}$. Although not defined in the algorithm, these may also be scaled by the number of instances in the training set.

Large positive values in the contribution suggest the original value of the feature, permuted in $v^{\prime}$, contributed strongly to $v$ being in the decided class. Small positive values suggest the feature weakly contributed to $v$ being in the decided class. Large negative values suggest that the original value of the feature that was permuted in $v^{\prime}$ strongly moved $v$ away from the decided class. Small negative values suggest that the feature weakly moved $v$ away from the decided class. Values of zero in the contribution suggest that the feature did not contribute to the decision-making process. 

Implicit in the calculation of contribution is also one of closeness. Closeness is the position in decision space of $v^{\prime}$ relative to members of the in-group and out-group of $v$. A feature may change how close $v$ is to other members of the in-group and out-group. This can be calculated as in Equation \ref{eq:average_proximity}. The proximity distances, however, are between $v$ and each $v^{\prime}$ for each member of the class (either in-group or out-group). This determines how each feature contributed to the instance $v$ being nearer or further from the in-group or out-group.

\section{Application of Feature Contribution and Closeness}

Feature contribution and closeness provide a new perspective on how data is used by the classifier. As such, we will show three main uses: 1) Demonstration of the decision-making process; 2) Explication of classification error; and 3) Inspection and identification of outliers in decision space.

\subsection{Classifier audit: Chemical structures for use in biofuels}

Prior work on chemical activity prediction [\cite{whitmore2016biocompoundml}] and structural interpretation [\cite{whitmore2016mapping}] showed how Random Forest classifiers could be used with LIME [\cite{ribeiro2016model}] to elucidate the contributions of molecular structural to the classification of chemicals in terms of research octane number. Assessing the potential octane performance of novel chemicals is an important roadblock in novel biofuel adoption. In addition to research octane number, cetane is used in diesel and diesel-like fuels as a measure of chemical performance. Here we use a strictly structural Random Forest model, trained on 361 compounds to classify compounds as having either a high ($> 40$) or low ($<= 40$) cetane number, relative to the North American standard for diesel fuels. Because cetane is principally a suitable vs. unsuitable designator, this problem is a classification problem. Additionally, the prior research octane classifier ($>94.4$ vs $<= 94.4$) classification is also examined.

Using a 6135 solely structural features (i.e., based on bond and atom presence vs. absence), a simple cetane Random Forest model has 83.04\% accuracy averaged over 100 permutations with 50\% randomized holdout and 85.33\% accuracy with 10x cross validation. This Random Forest cetane classifier allows rapid prediction and ranking of novel chemicals in terms of potential performance in diesel-like engines. The previously reported research octane classifier, performs with 85.89\% accuracy over 100 permutations with 50\% randomized holdout and 86.39\% accuracy with 10x cross validation.

One of the challenges to adoption of machine learning in biofuels research has been demonstrating the intuition behind the decision-making. The economic motivation prefers more time on model building to reduce risk during the expensive testing and roll-out phase. In this analysis we evaluate cetane and research octane number classification of these chemicals. These fuel characteristics are at a fundamental level, conversely related - albeit measured in independent ways. Cetane measures the propensity of a chemical to ignite under compression (the reason for it being used as a quality measure for diesel engines), whereas research octane number measures a chemical's resistance to pre-ignition (the reason for it being used as a quality measure for spark ignition engines). Good intuition for the suitability of a classifier would find that the presence of structural characteristics in a molecule would contribute in opposite ways to the high cetane and high research octane number. Since the models are using the same features, but not sharing decisions, robust models should come to the same conclusions about the relationships between structure and combustion in converse directions.

\begin{figure}[t]
\begin{center}
\centerline{\includegraphics[width=\columnwidth]{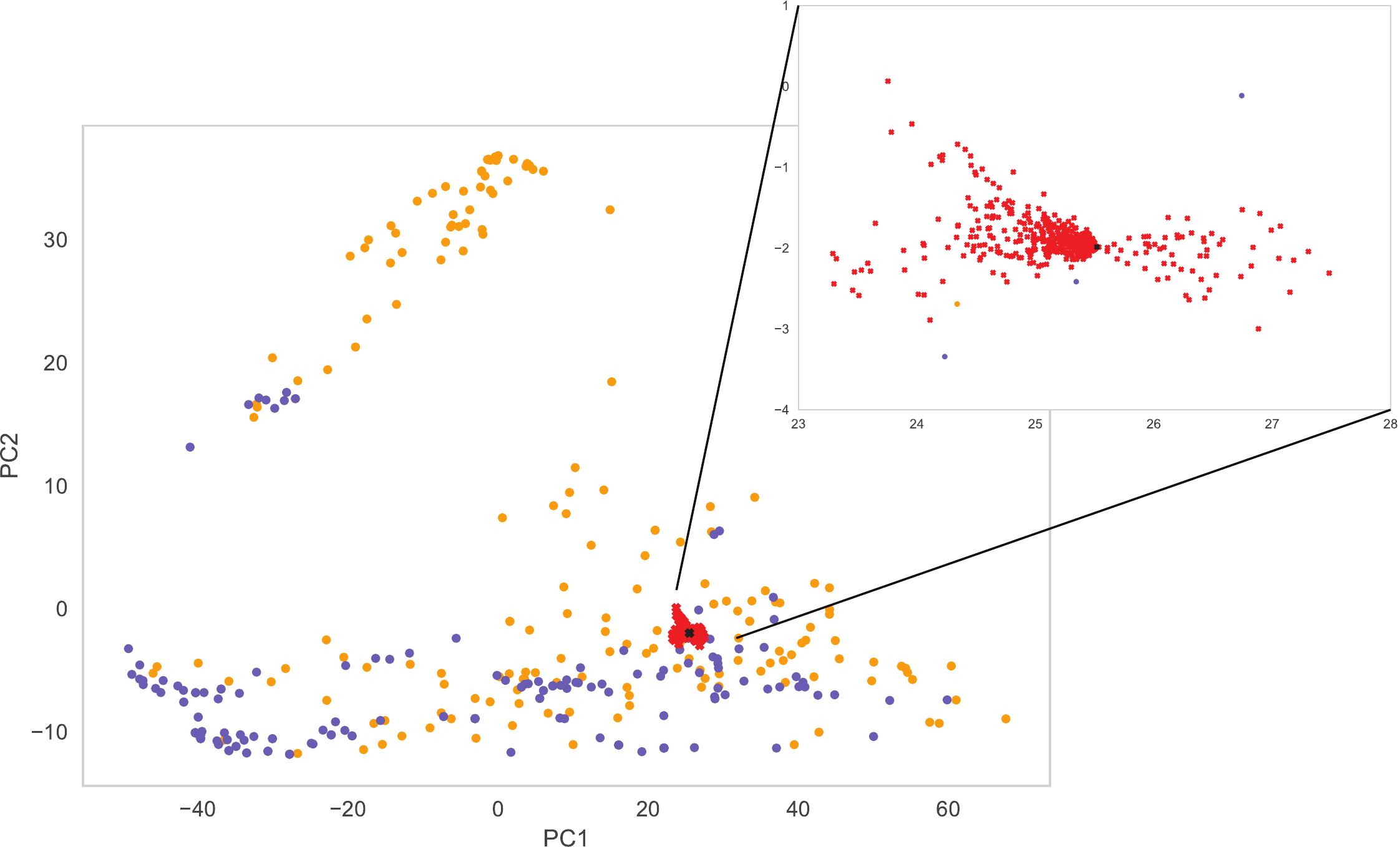}}
\caption{Low-dimensional proximity distance space (mapped onto the first two Principal Components) for 2-Methyltetrahydrofuran in cetane model. Orange dots are predicted low cetane and purple dots are predicted high cetane. The black dot is 2-Methyltetrahydrofuran. The red area corresponds to permuting individual features and the resulting change in space of the instance.}
\label{fig:cn_prox}
\end{center}
\vskip -0.2in
\end{figure}

Using the above proximity distance techniques it is possible to determine how changing a structural feature influences the decision-making process (see Fig \ref{fig:cn_prox}). For instance, removing or adding a particular part of the molecule that has been classified as high cetane, may lead to the molecule moving closer or further from the other molecules that have been similarly classified.

\begin{figure}[t]
\begin{center}
\centerline{\includegraphics[width=\columnwidth]{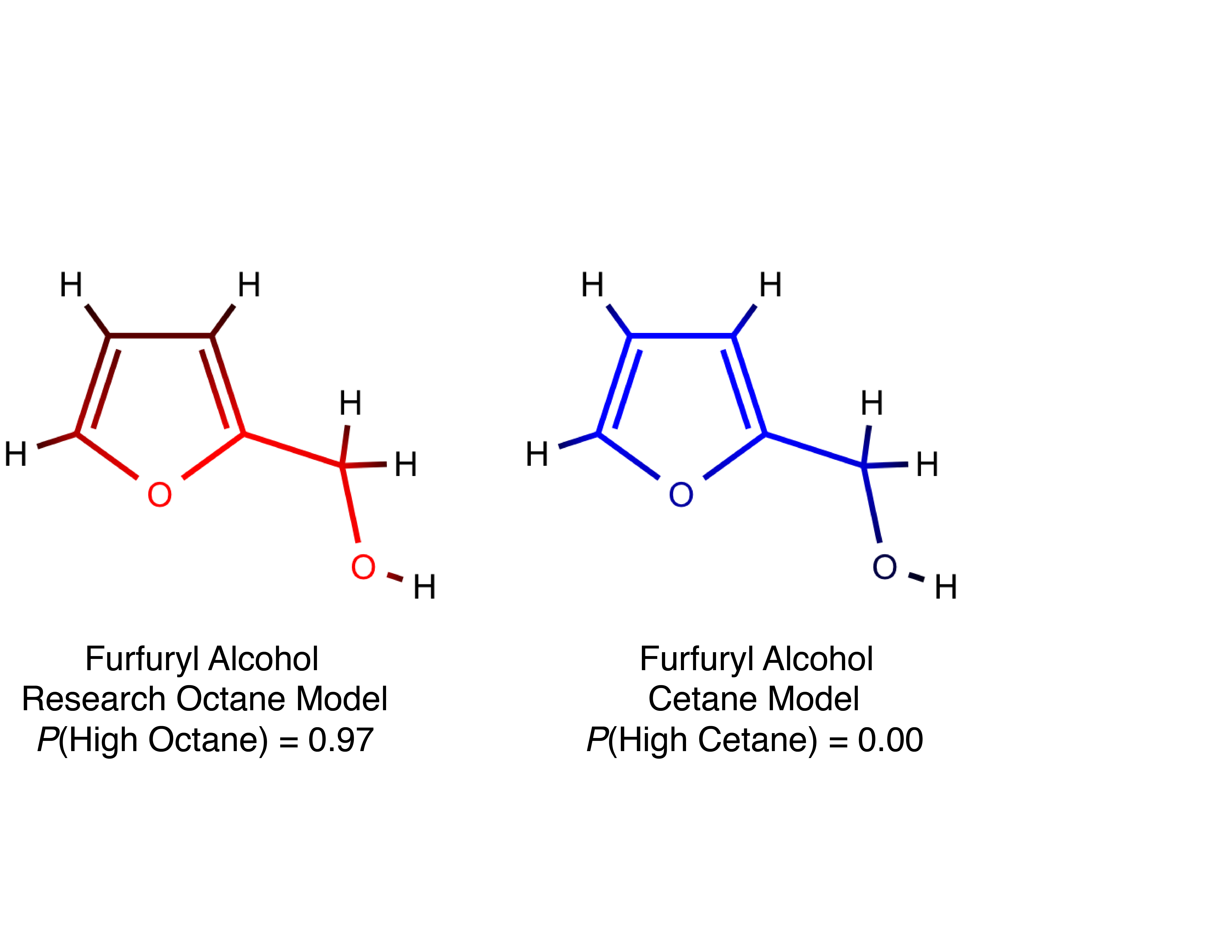}}
\caption{Mapping structural characteristics onto chemical structures. Furfuryl alcohol is a by-product of the degradation of organic products (e.g., wood, corncobs, switchgrass). This mapping varies from bright blue to bright red depending on how much each part of the molecule contributed to the high valued decision. For the research octane model (left), the bright portions contributed positively to a high octane decision. For the cetane model (right), the bright blue sections contributed negatively to a high cetane decision. The research octane model predicted that furfuryl alcohol is high octane, the cetane model predicted that it is low cetane. The mapping of individual features suggests a negative relationship between the relationships between structure and the research octane and cetane models.}
\label{fig:furfuryl}
\end{center}
\vskip -0.2in
\end{figure}

For molecules, positive and negative contributions from in-group and out-group proximity distance for particular atoms and bonds can be remapped back onto the molecular 2-dimensional structure (see Fig \ref{fig:furfuryl}). These maps highlight the projection of feature space, onto decision space. Each permutation in these bonds has the effect of moving the outcome nearer or further from other molecules in the in-group. 

Since the underlying process in the research octane number and cetane number models is assumed to be a mapping of chemical structures to combustion characteristics, it is expected that the carry-over from the feature contributions should be jointly informative in an audit of both models. To test this, we used the entire training set for both models, 428 chemicals and determined all feature contributions for both models. In instances where at least one prediction was high (either high research octane or high cetane), all non-zero valued features were collected. The hypothesis being tested is that feature contributions between the two models should be inversely correlated, such that a negative value for a feature contribution in predicting a high value in one model (e.g., high research octane) corresponds to a positive value for the same feature in predicting the high value in the other model (e.g., high cetane). 

\begin{table}[htbp]
  	\caption{Comparison of feature values in research octane and cetane classifiers, counts and (Pearson's Residuals)}
    \label{tab:table1}
    \centering
    \begin{tabular}{lrrrr}
         &                      \multicolumn{2}{c}{Research octane}                      &                           \\
    \cline{2-3}
    Cetane  &\multicolumn{1}{c}{Negative}&\multicolumn{1}{c}{Positive}&\multicolumn{1}{c}{Total}\\
    \midrule
    Negative   &             13875        &             61634        &              75509       \\
               &        (-62.43)          &       (41.88)        \\
    Positive    &             25338        &             25513        &             50851       \\
                &       (76.08)        &       (-51.04)        \\
    \midrule
    Total       &         39213        &             87147        &             126360         \\
    \bottomrule
    \end{tabular}
\end{table}

For the 428 chemicals, there were 126,360 non-zero ( i.e., $\mid x \mid >  0.0001$) features, where the chemical was predicted to be high valued in at least one of the models (this avoids the problem of cases where neither model was high valued, which may be the case for molecules that simply fail to combust). The correlation between paired feature values between these two models is negative and significant (Pearson's $r = -0.168$, Spearman's $\rho = -0.362$, Kendell's $\tau=-0.264$, $P = 0.0$). The correlation is subject to rank issues in the dataset. Taking a simple comparison of positives vs. negatives gives similar results (see Table \ref{tab:table1}). The $\chi^{2}$ statistic 14045.57 for df = 1 is significant (P = 0.0). The Pearson's Residuals are positive for (negative vs. positive) and (positive vs. negative) contingencies and negative for (positive vs. positive) and (negative vs. negative) contingencies, suggesting that the data is dominated by inverse relationships, suggesting that a feature that positively contributes to high research octane prediction, negatively contributes to high cetane prediction, and vice versa.

\subsection{Classification error: Post-hoc analysis of MNIST errors}

Another justification for explicable models, is understanding the failure modes. Oftentimes, even well performing models fail in ways that may be difficult to interpret or to diagnose. In this example we use 120,000 MNIST numerical figures, split between 60,000 training and six sets of 10,000 test instances. For a Random Forest, run using 1000 trees, after binary discretization of the images, the training accuracy was 100\% and the testing accuracy was 97.05\%. This means that 295 instances were misclassified in the testing data. Understanding why those instances were misclassified can be done using the proximity distance permutation listed above.

For a given misclassified instance, certain features contribute to the correct classification and the incorrect classification. This can be directly assessed by permuting each feature. For binary transformed MNIST data, this means bit-flipping individual pixels and determining how each one moves the instance nearer or further from the in-group.

\begin{figure}[t]
\begin{center}
\centerline{\includegraphics[width=\columnwidth]{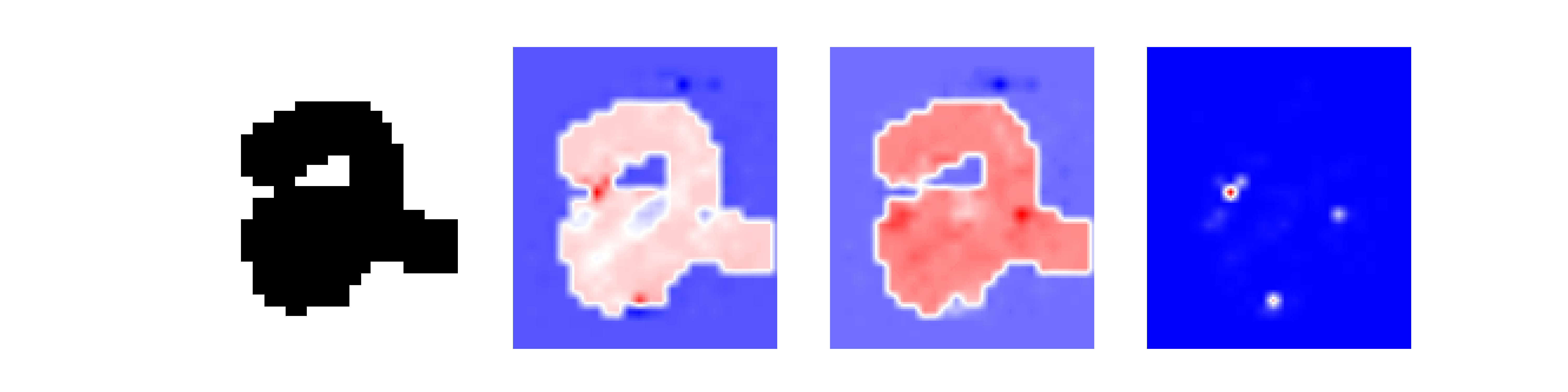}}
\caption{MNIST misclassification instance. Left panel. This is the binary representation of a two in the MNIST dataset, which was misclassified as an eight. Left center panel. This is a representation of the feature contribution for the incorrect classification. Dark red values indicate pixels that strongly weighted toward misclassification. Right center panel. This is a representation of the feature contributions for the correct classification. Dark red values indicate pixels that strongly weighted the model toward correct classification. Right panel. This is the squared difference between the two center panels. Dark red pixels correspond to differences necessary to convert between the misclassification and a correct classification.}
\label{fig:misclassified}
\end{center}
\vskip -0.2in
\end{figure}

Figure \ref{fig:misclassified} illustrates an example of this. Each pixel contributes to the correct classification in a negative or positive way, by moving the instance nearer or further from other instances in the same classification. Likewise, each pixel contributes to the incorrect classification in the same manner. The difference between misclassified instance and a correctly classified instance, is that the misclassified instances has some features that overwhelm the decision-making process, pointing to the incorrect classification. The right panel in Fig \ref{fig:misclassified} is a squared difference between the two feature contributions, and amounts to a permutation matrix, quantifying the transformation necessary to convert between incorrect and correct classifications.

\section{Conclusion}

By combining the intuition behind tools like LIME, which directly manipulate input variables and observe changes in output, with decision, rather than feature space, we have developed a tool that provides robust explication of the decision-making process. Unlike LIME, this technique is not machine agnostic, but rather specific to Random Forests. Tools like path counting and mean decrease in accuracy are notably simpler then the technique we have presented, but limit the resulting information. The proximity method has two outcome measures: 1) contribution of a feature to its prediction, and 2) closeness that a feature contributes to the in-group class membership. Both of these outcomes are important in determining how valuable a feature is in correctly or incorrectly predicting the outcome value.

Random Forests and their associated algorithms are a very commonly used set of machine learning tools. Proximity has a long history as a technique for quantifying the Random Forest decision space. We present its use in unraveling Random Forests in a way not previously described. This method is not a standard tool for sensitivity analysis in Random Forests or machine learning. The goal here is to connect a standard technique for permutation of input to the topological representation of decision space. This technique is more nuanced than changing input values and observing changes in prediction, providing users with hypothetical relationships to outcome, mapped on feature values. Three important characteristics of explainable artificial intelligence are 1) The ability to represent the individual decision-making process in human terms; 2) The ability to determine the bounds on the correctness of the algorithm, without relying simply on the measures of accuracy and error that the algorithm originally optimized; and 3) The ability to analyze decision-making errors to assess the modes of failure. Improving these three areas is an important step in developing robust, safe and responsible artificial intelligence. This work will hopefully extend intervention analysis and provide more causal explanations of data, which will lead to more useful outcomes.

\subsubsection*{Acknowledgments}

We thank David W. Aha and the anonymous reviewers of the “IJCAI/ECAI 2018 Workshop on Explainable Artificial Intelligence (XAI)” for their very useful comments and suggestions.\\

Sandia National Laboratories is a multimission laboratory managed and operated by National Technology and Engineering Solutions of Sandia, LLC., a wholly owned subsidiary of Honeywell International, Inc., for the U.S. Department of Energy’s National Nuclear Security Administration under contract DE-NA0003525.\\

This research was conducted as part of the Co-Optimization of Fuels \& Engines (Co-Optima) project sponsored by the U.S. Department of Energy (DOE) Office of Energy Efficiency and Renewable Energy (EERE), Bioenergy Technologies and Vehicle Technologies Offices. Co-Optima is a collaborative project of multiple national laboratories initiated to simultaneously accelerate the introduction of affordable, scalable, and sustainable biofuels and high-efficiency, low-emission vehicle engines.

\end{document}